%% file: main.tex
\documentclass{article}

\usepackage{microtype}
\usepackage{graphicx}
\usepackage{subfigure}
\usepackage{booktabs} 

\usepackage{hyperref}

\usepackage[accepted]{icml2024}

\usepackage{amsmath}
\usepackage{amssymb}
\usepackage{mathtools}
\usepackage{amsthm}

\usepackage[capitalize,noabbrev]{cleveref}

\theoremstyle{plain}

\theoremstyle{definition}

\theoremstyle{remark}

\usepackage[textsize=tiny]{todonotes}

\usepackage{amsmath}
\usepackage{amssymb}
\usepackage{textcomp}
\usepackage{bm}
\usepackage{pifont}
\usepackage{tcolorbox}
\tcbuselibrary{skins}
\usepackage{multirow}
\usepackage{booktabs} 
\usepackage{tikz}
\usepackage{wrapfig}
\usepackage{xcolor}

\usepackage{enumitem}

\icmltitlerunning{Beyond the Federation: Topology-aware Federated Learning for Generalization to Unseen Clients}

\begin{document}

\twocolumn[
\icmltitle{Beyond the Federation: Topology-aware Federated Learning for\\ Generalization to Unseen Clients}



\icmlsetsymbol{equal}{*}

\begin{icmlauthorlist}
\icmlauthor{Mengmeng Ma}{ud}
\icmlauthor{Tang Li}{ud}
\icmlauthor{Xi Peng}{ud}
\end{icmlauthorlist}

\icmlaffiliation{ud}{DeepREAL Lab, Department of Computer \& Information Sciences, University of Delaware}

\icmlcorrespondingauthor{Mengmeng Ma}{mengma@udel.edu}

\icmlkeywords{Federated Learning, Out-of-federation Generalization}

\vskip 0.3in
]



\printAffiliationsAndNotice{}  

\begin{abstract}
Federated Learning is widely employed to tackle distributed sensitive data. Existing methods primarily focus on addressing in-federation data heterogeneity. However, we observed that they suffer from significant performance degradation when applied to unseen clients for out-of-federation (OOF) generalization. The recent attempts to address generalization to unseen clients generally struggle to scale up to large-scale distributed settings due to high communication or computation costs. Moreover, methods that scale well often demonstrate poor generalization capability. To achieve OOF-resiliency in a scalable manner, we propose Topology-aware Federated Learning (TFL) that leverages client topology - a graph representing client relationships - to effectively train robust models against OOF data. We formulate a novel optimization problem for TFL, consisting of two key modules: Client Topology Learning, which infers the client relationships in a privacy-preserving manner, and Learning on Client Topology, which leverages the learned topology to identify influential clients and harness this information into the FL optimization process to efficiently build robust models. Empirical evaluation on a variety of real-world datasets verifies TFL's superior OOF robustness and scalability. 
\end{abstract}


\input{sec_submit/sec1_introduction}
\input{sec_submit/sec2_prelim}
\input{sec_submit/sec3_method}

\input{sec_submit/sec4_experiments}

\input{sec_submit/sec5_releated}
\input{sec_submit/sec6_conclusion}
\input{sec_submit/sec7_impack}


\bibliography{refrences}
\bibliographystyle{icml2024}

\input{sec8_supps}

\end{document}

%% file: sec_submit/sec1_introduction.tex
\section{Introduction}
\begin{figure}
    \centering
    \includegraphics[trim=0cm 0cm 0cm 0.0cm,clip, width=0.42\textwidth]{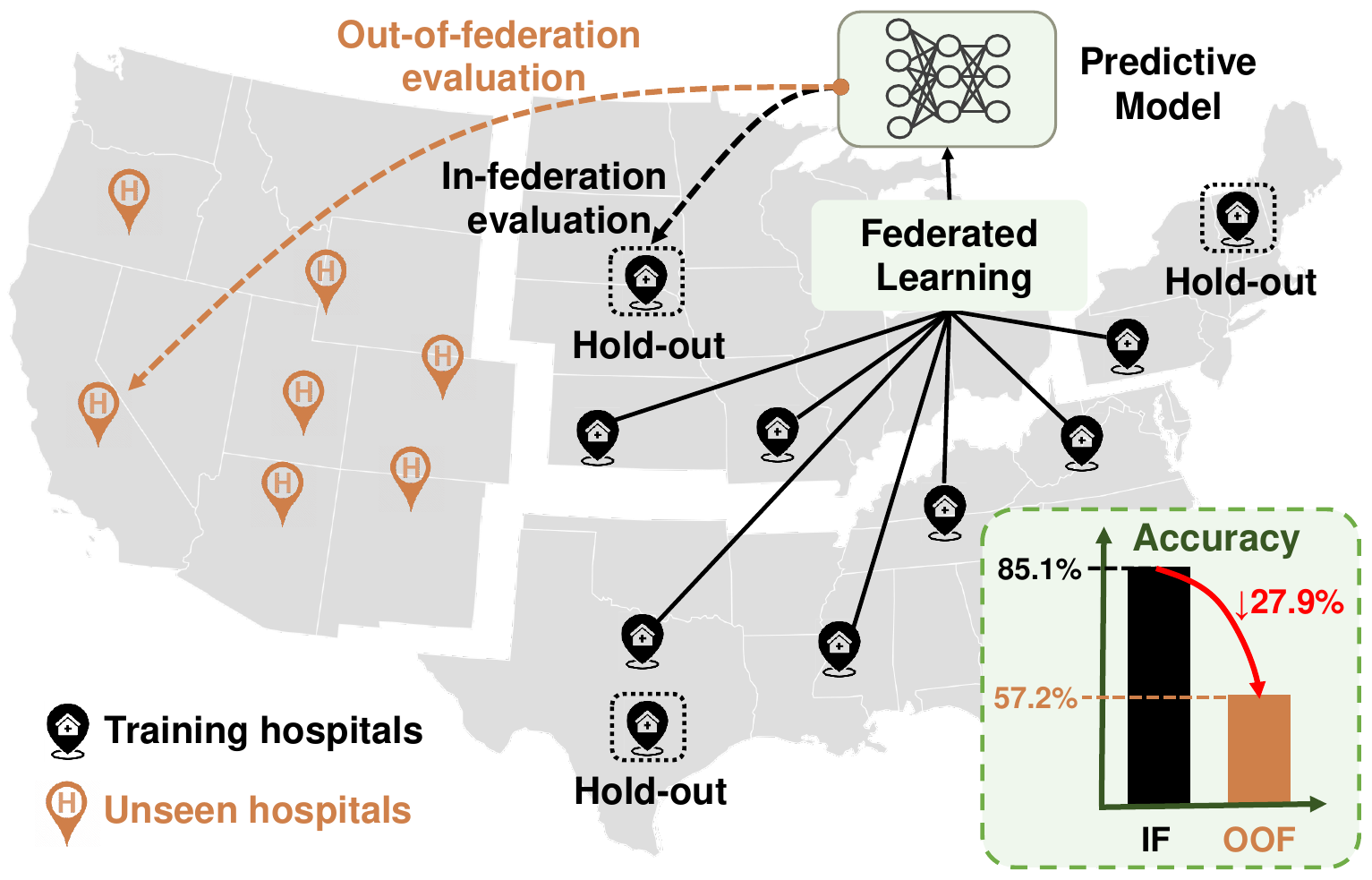}
    \vspace{-7pt}
    \caption{Our empirical evaluation of patient mortality prediction using Federated Learning on distributed Healthcare dataset (eICU~\citep{pollard2018eicu}). We observed that \textit{a model exhibiting high accuracy with in-federation (IF) data can fail catastrophically when presented with out-of-federation (OOF) data.}}
    \label{fig:ood}
    \vskip -0.2in
\end{figure}

Federated Learning (FL) has emerged as a promising solution for handling distributed sensitive data, enabling multi-institutional collaboration by distributing model training to data owners and aggregating results on a centralized server~\cite{mcmahan2017communication}. 
This data-decentralized approach harnesses the collective intelligence of all participating nodes to build a model that is potentially more robust and generalizable~\cite{sheller2020federated}. Existing robust FL methods~\cite{li2020federated,deng2020distributionally} primarily focus on learning a global model with good average or worst-case performance, addressing \textit{in-federation (IF)} data heterogeneity. 
However, these methods can fail catastrophically on \textit{out-of-federation (OOF)} clients, \textit{i.e.,} clients outside the collaborative federation. The OOF clients pose significant generalization challenges, as FL models may encounter \textit{unseen} distributions outside their training space~\cite{pati2022federated}. Our empirical study shows that existing methods suffer from significant performance degradation when applied to unseen clients for OOF generalization (see Figure~\ref{fig:ood}).

\begin{figure}
    \centering
    \includegraphics[trim=0cm 0cm 0cm 0.0cm,clip, width=0.45\textwidth]{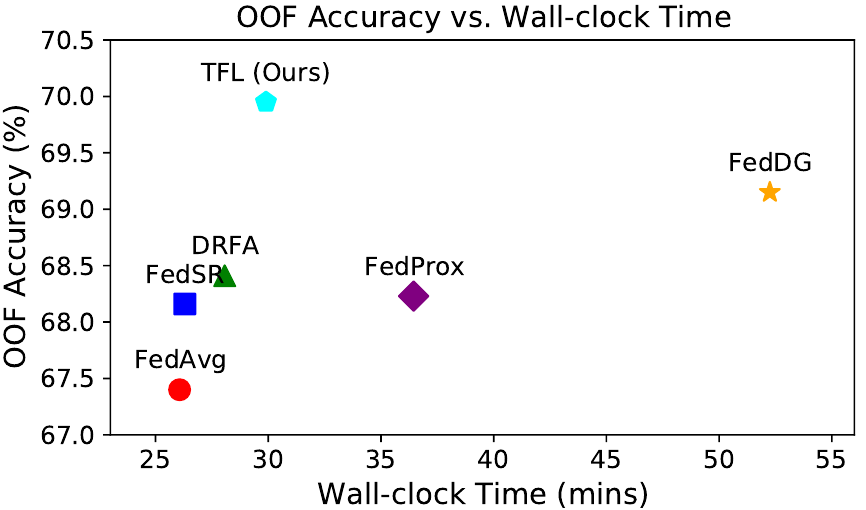}
    \vspace{-10pt}
    \caption{Accuracy vs. wall-clock time on PACS dataset~\cite{li2017pacs}. Wall-clock time is used as a holistic evaluation of scalability (communication and computation costs). \textit{We see a tradeoff between OOF robustness and scalability in existing methods.}}
  \label{fig:tradeoff}
  \vskip -0.25in
\end{figure}

There have been recent attempts to address the challenge of unseen clients through client augmentation and client alignment. However, these approaches encounter difficulties in large-scale distributed environments. Client augmentation methods~\cite{liu2021feddg} often necessitate intensive client-server communication, leading to a notable \textit{communication bottleneck} that hinders scalability~\cite{zhou2023efficient}. Client alignment methods~\cite{zhang2021federated}, which operate by aligning client distributions through adversarial training, are less communication costly but introduce substantial \textit{computation burdens} due to complex local training process~\cite{bai2023benchmarking}. While using a fixed reference distribution for alignment~\cite{nguyen2022fedsr} offers computational efficiency, it has shown \textit{limited generalization capabilities to unseen clients}~\cite{bai2023benchmarking}. Our empirical evaluation (Figure~\ref{fig:tradeoff}) reveals the tradeoff between OOF resiliency and scalability in current methods. A question naturally arises: Can we build OOF robust models in a scalable manner?  

We propose to trigger OOF-resiliency while being scalable by leveraging \textit{client topology}. Client topology, a graph representing client relationships, allows for using graph mining techniques~\cite{saxena2020centrality,lu2011link} to derive insights into client data distributions. It can be used to identify ``influential'' clients that are representative of the training clients, containing distributions more likely to be encountered in OOF clients. 
For instance, an influential client could be a regional hospital that aggregates a diverse range of patient data. This hospital's data encapsulates a rich repository of information, mirroring the variety and complexity of data that could be seen in OOF scenarios. Leveraging these influential clients as priority contributors in the training rounds can facilitate the model in learning from the most representative data, thereby potentially enhancing its OOF robustness. On the other hand, by reducing unnecessary communication with non-influential clients, communication costs can be significantly reduced. 

Grounded on the concept of client topology, we formulate the problem of generalization to unseen clients as the optimization of a joint objective over the models and the client topology. To solve this optimization problem, we propose Topology-aware Federated Learning, which consists of two steps: 1) {Client Topology Learning}: Inferring the client topology while respecting data privacy. 2) {Learning on Client Topology}: Leveraging the learned topology to build a robust model. The first step learn a client topology that reflects client relationships by analyzing pairwise model similarity. In the second step, a robust model is efficiently optimized by harnessing the client's influential information to regularize a distributed robust optimization process.

Our contributions are as follows: Firstly, we introduce the \textit{Topology-aware Federated Learning} (TFL), a scalable framework designed to enhance FL's out-of-federation (OOF) robustness. TFL utilizes client relationships to develop robust models against OOF data.
Secondly, we design an iterative client topology learning and learning on client topology approach to solve TFL. Finally, we have \textit{curated two OOF benchmarks} using real-world healthcare data, offering valuable testbeds for subsequent research. Through extensive experiments on curated and standard benchmarks, we verify TFL's superior OOF-resiliency and scalability.

%% file: sec_submit/sec2_prelim.tex
\section{Preliminaries}
We start by presenting the average-case formulation of FL and subsequently introduce a fundamental robust optimization framework for FL (worst-case formulation).

\begin{figure*}[t]
\begin{center}
\includegraphics[width=0.93\linewidth]{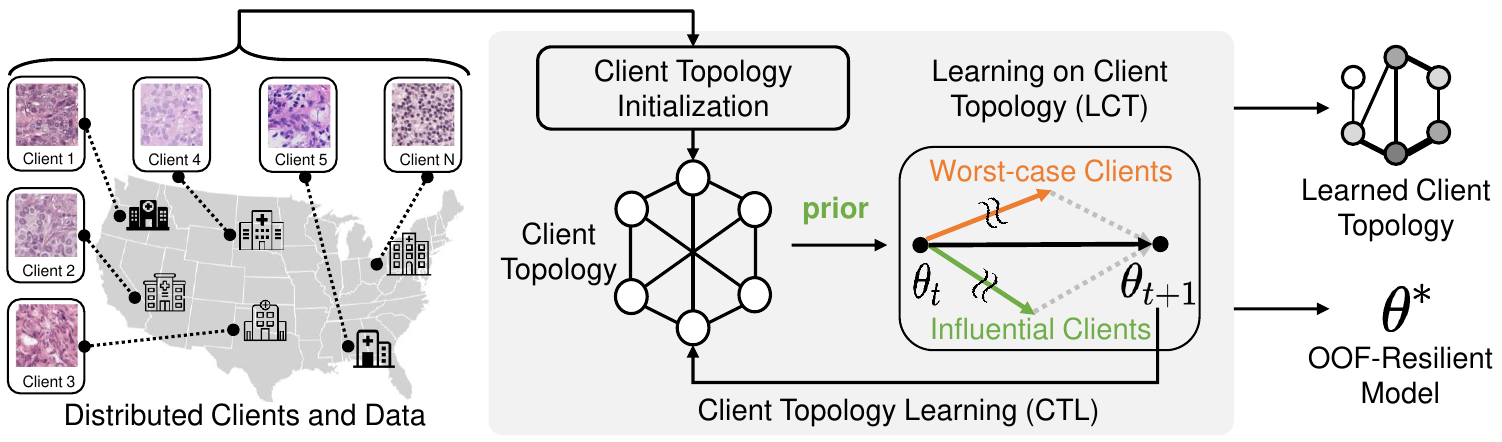}
\end{center}
\caption{Overview of Topology-aware Federated Learning (TFL). TFL contains two alternating steps: \textit{client topology learning} (CTL) and \textit{learning on client topology} (LCT). \textbf{CTL} learn the client topology that describes the relationships between local clients. 
We use model weights as node embedding and construct a $\epsilon$-graph by measuring the similarity of node pairs. 
\textbf{LCT} leverage the learned client topology to achieve better OOF robustness. We identify the influential client and then use the influential client as prior knowledge to regularize a distributionally robust optimization framework. 
In this way, the optimization process can balance the ``worst-case'' client and the ``influential'' client to avoid overly pessimistic models with compromised OOF-resiliency.}
\vskip -0.15in
\label{fig:overview}
\end{figure*}

\textbf{Federated learning (Average-case formulation).} The standard FL involves collaboratively training a global model leveraging data distributed at $K$ clients. 
Each client $k$ ($1 \leq k \leq K$) has its own data distribution $D_k (x, y)$, where $x \in \mathcal{X} $ is the input and $y \in \mathcal{Y}$ is the label, and a dataset with $n_k$ data points: $\widehat{D}_k= \left \{  (x_k^n, y_k^n) \right \}_{n=1}^{n_k}$. Local data distributions $\left \{ D_k \right \}_{k=1}^K$ could be the same or different across the clients. FL aims to learn a global model $\theta$ by minimizing the following objective function:
\vspace{-11pt}
\begin{equation}
\label{eqn:fedavg}
 \min_{\theta \in \Theta}\ F(\theta), \; \text{where}\;  F(\theta) := \sum_{k=1}^{K} p_k f_k(\theta),
\end{equation}
where $f_k(\theta)$ is the local objective function of client $k$. The local objective function is often defined as the empirical risk over local data, \textit{i.e.,} $f_k (\theta) = \mathbb{E}_{(x,y) \sim \widehat{D}_k} [{\ell(\theta;x,y)}] = \frac{1}{n_k} \sum_{n=1}^{n_k} \ell(\theta;x_k^{n}, y_k^{n})$. The term $p_k$ ($p_k \geq 0$ and $\sum_k p_k = 1$) specifies the relative importance of each client, with two settings being $p_k = \frac{1}{N}$ or $p_k = \frac{n_k}{N}$, where $N = \sum_{k} n_k$ is the total number of samples.

\textbf{Distributionally robust federated learning (Worst-case formulation).} While Equation~\ref{eqn:fedavg} can build a global model with good \textit{average performance} on in-federation clients, it may not necessarily guarantee good performance in the presence of heterogeneous data. In the real world, data could be statistically heterogeneous due to different data acquisition protocols or various local demographics~\cite{rieke2020future}. Thus, the local data distributions may deviate significantly from the average distribution, implying that an ``average'' global model can fail catastrophically under distributional drift.
To tackle this issue, distributionally robust optimization~\cite{staib2019distributionally} has been adapted to FL, resulting in distributionally robust federated learning~\cite{deng2020distributionally,reisizadeh2020robust}. The formulation of this new problem is as follows:
\vspace{-5pt}
\begin{equation}
\label{eqn:drfl}
\begin{aligned}
\min_{\theta \in \Theta}\, \max_{\bm{\lambda} \in \Delta_K}\, & F(\theta , \bm{\lambda}) := \sum_{k=1}^{K} \lambda_k f_k(\theta),
\end{aligned}
\end{equation}
where $\bm{\lambda}$ is the global weight for each local loss function and $\Delta_K$ denotes the $K-1$ probability simplex. Intuitively, Equation~\ref{eqn:drfl} minimizes the maximal risk over the combination of empirical local distributions, and therefore the worst-case clients would be prioritized during training.

While this framework has the potential to address distribution shifts~\cite{mohri2019agnostic,qiao2023topology,deng2020distributionally}, directly implementing it for OOF resiliency may yield suboptimal models. This approach heavily relies on \textit{worst-case clients}, those with large empirical risks, to develop robust models. However, these clients may not necessarily be the \textit{influential ones} that are representative of clients. In some cases, it is possible that this approach overly focuses on ``outlier'' clients, clients that are significantly different from most of the training clients, leading to models with limited OOF robustness. Therefore, we argue that, to build optimal OOF-resilient models, the optimization process should focus on not only the worst-case but also the influential clients. 

%% file: sec_submit/sec3_method.tex
\section{Methodology}

In this section, we introduce the proposed \textit{Topology-aware Federated Learning} (TFL) framework. TFL aims to leverage client topology to improve the model's OOF robustness.
We use graphs as tools to describe the client topology because they provide a flexible way to represent the relationship between distributed clients~\cite{dong2019learning,mateos2019connecting}. Specifically, we model client topology using an undirected graph~\cite{vanhaesebrouck2017decentralized,ye2023personalized}. In the graph, nodes correspond to clients, and edges reflect client connectivity. Let $\mathcal{G} =(V, E, W)$ denote the client topology, where $V$ is the node set with $\left | V \right | = K$, $E \subseteq V \times V $ is the edge set and $W \in \mathbb{R}^{K \times K}$ is the adjacency matrix.  An edge between two clients $v_k$ and $v_l$ is represented by $e_{k,l}$ and is associated with a weight $w_{k,l}$. 

\textbf{Optimization problem.} 
Let $\phi$ denotes graph measures, \textit{e.g,} centrality measure~\citep{saxena2020centrality}. The function ``sim'' indicates any similarity function, including but not limited to cosine similarity, $\ell_2$, or $\ell_1$. $\gamma$ represents a trade-off hyperparameter. $\mathcal{D}$ symbolizes arbitrary distributional distance measure. As the client topology is often not readily available, we propose to jointly optimize the client topology and the robust model by solving the following problem:
\vspace{-5pt}
\begin{equation}
\begin{aligned}
\min_{\substack{\theta \in \Theta \\ w \in W }}\, \max_{\bm{\lambda} \in \Delta_K}\,  & F(\theta, \bm{\lambda}, W) :=  \sum_{k=1}^{K} \lambda_k f_k(\theta) \\  & \quad\quad - \frac{\gamma}{2} \sum_{k \neq l} w_{k,l} \; \text{sim}(v_k, v_l) , \\
 \textrm{s.t.}\;\; & \mathcal{D}(\bm{\lambda} \parallel \bm{p}) \leq \tau,\, \text{where}\; \bm{p} = \phi(W). 
\end{aligned}
\label{eqn:tfl}
\end{equation}
In the proposed objective function, the first term follows the same spirit of Equation~\ref{eqn:drfl} to adopt a minimax robust optimization framework. The difference is that it minimizes the risk over not only the worst-case but also the influential clients. The second term is dedicated to learning the client topology by measuring the pair-wise client similarity.


Our formulation stands apart from existing work in two respects. First, it employs client topology to explicitly model the client relationships. Analyzing this topology facilitates the identification of influential clients that are crucial for developing OOF generalization. Second, our formulation enables seamless integration of client topology into the optimization process, ensuring that the model assimilates insights from the most significant clients.

In Equation~\ref{eqn:tfl}, it is infeasible to simultaneously update both the client topology $W$ and model parameters $\theta$ as local clients do not have access to other clients' data.
To this end, we propose to solve this problem using an alternating two-step approach: learning the client topology (updating $W$) and building an OOF-resilient model with client topology (updating $\lambda$ and $\theta$). As shown in Figure~\ref{fig:overview}, we term these steps as \textbf{Client Topology Learning} (CTL) and \textbf{Learning on Client Topology} (LCT), respectively. The following sections will provide further details on these two steps.


\subsection{Client Topology Learning}
\label{subsec:3.1}

Our goal is to learn the client topology that accurately describes the characteristics of local data, capturing the underlying relationships among clients. Conventional approaches typically adopt similarity-based~\citep{chen2020iterative,franceschi2019learning} or diffusion-based~\citep{zhu2021deep,mateos2019connecting} methods to estimate the graph structure from data. However, most methods require centralizing training data on a single machine, thereby raising privacy concerns. Therefore, the primary challenge lies in learning the client topology while respecting data privacy.

\begin{figure}[t]
    \centering
    \includegraphics[trim=0cm 0cm 0cm 0.0cm,clip, width=0.45\textwidth]{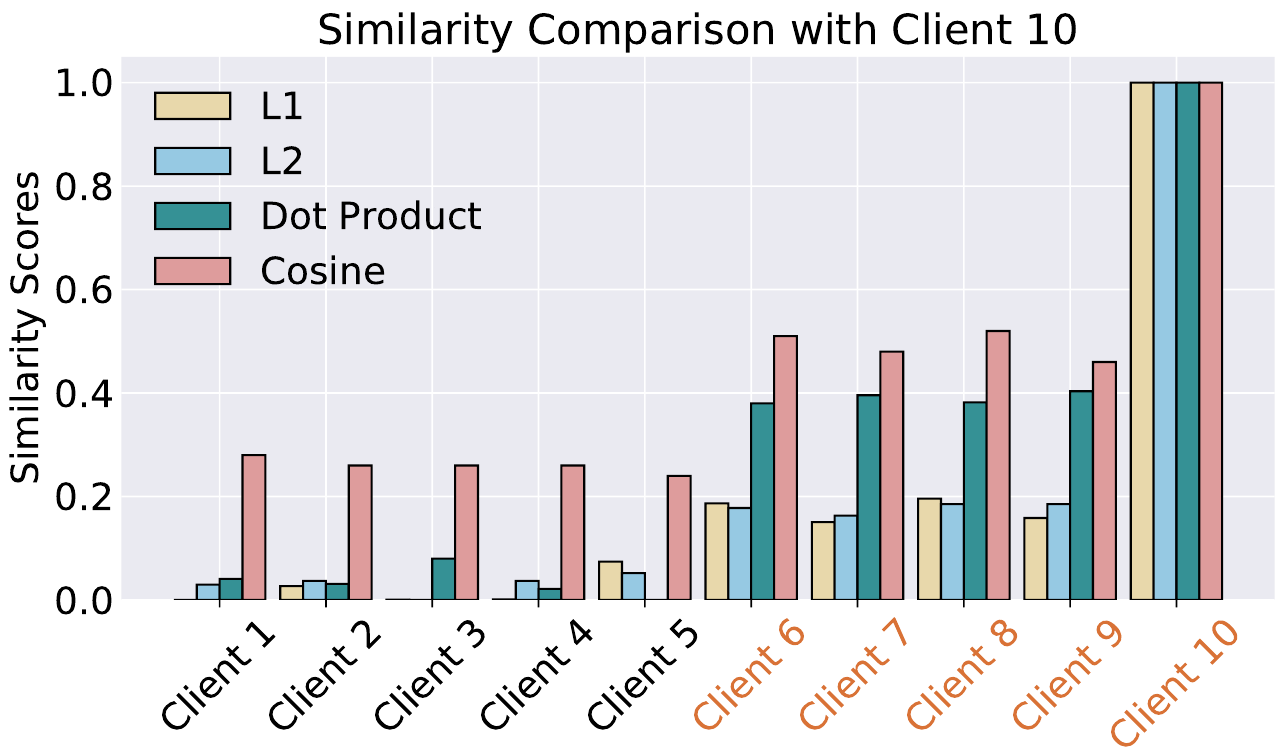}
    \caption{Model similarity under different metrics. Client models are trained using the same algorithm and hyperparameters. Clients 6 to 10 share similar data distributions. We observe that \textit{clients with similar data distribution tend to have more similar models}.}
  \label{fig:sim}
    \vskip -0.2in
\end{figure}

We propose to utilize model weights to learn client topology.
Intuitively, when two clients have similar data distributions, their corresponding models should be more similar. It is possible to obtain data distribution relationships with model similarity. While the feasibility of this idea is supported by literature~\citep{yu2022predicting}, we intend to empirically verify whether it holds in our setting. We use four types of similarity measures, including $\ell_1$-based, $\ell_2$-based, dot product-based, and cosine similarity-based metrics. We conduct experiments on PACS where clients 6 to 10 share similar data distributions. Then, we measure the similarity between client 10 and all other clients. For a clear comparison, we normalized all scores into $[0, 1]$. The results are shown in Figure~\ref{fig:sim}. We observe that clients sharing similar data distributions exhibit significantly higher similarity scores compared to others. Furthermore, these distinct similarity scores remain consistent across various metrics.

In summary, utilizing model weights to learn client topology offers two merits. First, models can be freely shared among clients, addressing privacy issues of client topology learning. Second, since models are trained on local data, their similarity measures the similarity between local data distributions. We formulate client topology learning as follows:
\begin{equation}
\label{eqn:ctl}
\begin{aligned}
\min_{w \in W}\, - \sum_{k \neq l} w_{k,l}\; \text{sim}(\theta_k, \theta_l) + \left \| W \right \|_0, \\
\end{aligned}
\end{equation}
where $\theta$ denotes the model of local clients. This objective ensures that similar clients have large edge weights. Our formulation is aligned with the common network homophily assumption that \textit{edges tend to connect similar nodes}~\citep{newman2018networks}. 
To avoid learning a fully connected graph with trivial edges, we enforced graph sparsity by penalizing the $l_0$-norm of the adjacency matrix $\left \| W \right \|_0$. We implement the sparsity term using an implicit method, \textit{i.e.,} hard thresholding on $W$ to construct $\epsilon$-graph~\citep{zhu2021deep}. Specifically, we mask out (\textit{i.e.,} set to zero) those elements in $W$ smaller than a non-negative threshold $\epsilon$.

\subsection{Learning on Client Topology} 
\label{subsec:lot}
The learned client topology captures the relation among local clients. We aim to leverage such relations to develop TFL for better OOF-resiliency. Recall that, to tackle distribution shift, distributionally robust federated learning (DRFL) assumes that the target distribution lies within an arbitrary mixture of training distributions: $\sum_{k=1}^{K} \lambda_k {D}_k$. DRFL builds OOF-resilient models by minimizing the \textit{worst-case} risk over an uncertainty set of possible target distribution $ Q := \{\sum_{k=1}^K \lambda_k {D}_k \mid \bm{\lambda} \in \Delta_K \}$. 
However, DRFL primarily emphasizes the worst-case distribution, potentially ignoring the \textit{influential} ones that are more representative of the training clients. This can lead to overly pessimistic models with compromised OOF resiliency~\citep{hu2018does}.

We leverage client topology to construct an uncertainty set that can better approximate the unseen distribution. Our insight is to optimize the model for both the worst-case and influential distributions. The key challenge is how to identify the influential distribution. Our idea is to use graph centrality as the criterion to choose influential distributions. \textit{Graph centrality} is widely used in social network analysis~\citep{newman2005measure} to identify the key person by measuring how much information propagates through each entity. We introduce \textit{client centrality} to identify influential ones, which can be calculated by graph measurements such as degree, closeness, and betweenness. Specifically, we first calculate the centrality of each client in $\mathcal{G}$ as the \textit{topological prior} $\bm{p}$. Then we use $\bm{p}$ to constraint the uncertainty set $Q$ by solving the following minimax optimization problem:  
\begin{equation}
\label{eqn:tafl}
\begin{aligned}
\min_{\theta \in \Theta}\, \max_{\bm{\lambda} \in \Delta_K}\, & F(\theta , \bm{\lambda}) := \sum_{k=1}^{K} \lambda_k f_k(\theta),\;\; \\
\textrm{s.t.}\;\;  & \mathcal{D}(\bm{\lambda} \parallel \bm{p}) \leq \tau. 
\end{aligned}
\end{equation}

The topological constraint directs the optimization to focus on worst-case and influential clients.
Notably, both FedAvg~\citep{mcmahan2017communication} (Equation~\ref{eqn:fedavg}) and DRFL (Equation~\ref{eqn:drfl}) can be viewed as special cases of TFL: When $\tau = 0$ and the prior $\bm{p}$ is a uniform distribution, Equation~\ref{eqn:tafl} minimizes the average risks over local clients, which is identical to FedAvg; When $\tau \rightarrow \infty$, Equation~\ref{eqn:tafl} only prioritizes the worst-case clients, resembling the DRFL approach.

The above optimization problem is typically nonconvex, and methods such as SGD cannot guarantee constraint satisfaction~\citep{robey2021model}. To tackle this issue, we leverage the Lagrange multiplier and KKT conditions~\citep{boyd2004convex} to convert it into unconstrained optimization:
\begin{equation}
\label{eqn:tafl_un}
\begin{aligned}
\min_{\theta \in \Theta}\, \max_{\bm{\lambda} \in \Delta_K}\, & F(\theta , \bm{\lambda}, \bm{p}) := \sum_{k=1}^{K} \lambda_k f_k(\theta) - q \mathcal{D}(\bm{\lambda} \parallel \bm{p}),\\
\end{aligned}
\end{equation}
where $q$ is the dual variable. We solve the primal-dual problem by alternating between gradient descent and ascent:
\begin{equation}
\label{eqn:update}
\begin{aligned}
\theta^{t+1} & = \theta^t  - \eta_{\theta}^t \nabla_\theta F(\theta, \bm{\lambda}),\;\; \\
\bm{\lambda}^{t+1} &= \mathcal{P}_{\Delta_K} (\bm{\lambda}^t  + \eta_{\lambda}^t \nabla_{\lambda} F(\theta, \bm{\lambda})),
\end{aligned}
\end{equation}
where $\eta^t$ is the step size. $\mathcal{P}_{\Delta_K}(\bm{\lambda})$ projects $\bm{\lambda}$ onto the simplex $\Delta_K$ for regularization. 

\textbf{Influential client.} We identify the influential clients by calculating betweenness centrality. Betweenness centrality measures how often a node is on the shortest path between two other nodes in the graph $\mathcal{G}$. It has been revealed that the high betweenness centrality nodes have more control over the graph as more information will pass through them~\citep{white1994betweenness}. The betweenness centrality of client $k$ is given by the expression of $c_k=\sum_{s\ne k \ne t \in [K] } \frac{\sigma_{st}(k)}{\sigma_{st}}$, where $\sigma_{st}$ is the total number of shortest path from node $s$ to node $t$ ($(s,t)$-paths) and $\sigma_{st}(k)$ is the number of $(s,t)$-paths that pass through node $k$. Then we apply softmax to normalize client centrality $c_k$ to obtain client topological prior $p_k = \text{exp}(c_k)/\sum_{k=1}^K \text{exp}(c_k)$.

\textbf{Discussion.} \textbf{Handling high-dimensional models.}  Large models are getting more attention in FL~\citep{zhuang2023foundation}. Its high dimensionality might raise computational concerns when calculating model similarity. For better computation efficiency, we can leverage only partial model parameters, \textit{e.g.,} the last few layers. We empirically show that using partial parameters does not significantly affect the OOF performance (see results in Table~\ref{tab:approx}). \textbf{Client topology learning in cross-device settings}. 
Client topology learning incurs $\mathcal{O}(N^2)$ computation complexity for $N$ clients. This quadratic complexity could be prohibitively expensive in cross-device FL that involves thousands or millions of clients. We argue that computation costs can be significantly reduced via client clustering~\citep{sattler2020clustered}. Our empirical evaluation (see results in Table~\ref{tab:c_clustering}) shows that client clustering significantly reduces computation costs.

%% file: sec_submit/sec4_experiments.tex
\section{Experiments}

\subsection{Datasets and Baselines}
\label{exp:data}

\begin{table*}[t]
\vskip -0.1in
\centering
\caption{Accuracy on the PACS. We conduct experiments using \textit{leave-one-domain-out}, meaning each domain serves as the evaluation domain in turn. Existing methods typically consider each domain as an individual client~\citep{liu2021feddg,nguyen2022fedsr}. To simulate a large-scale setting, we further divide each domain into $5$ subsets and treat each subset as a separate client. The total number of clients is $20$. The reported numbers are from three independent runs. \textit{Our method outperformed others across all tested settings}.}
\vskip 0.1in
\resizebox{0.9\textwidth}{!}{
\begin{tabular}{cccccccc}
\toprule
\multicolumn{2}{c}{\multirow{2}{*}{Models}}                                                  & \multirow{2}{*}{Backbone} & \multicolumn{5}{c}{PACS}                \\ \cmidrule{4-8} 
\multicolumn{2}{c}{}                                                                         &                           & A     & C     & P     & S     & Average \\ \midrule
\multirow{2}{*}{\begin{tabular}[c]{@{}c@{}}Centralized \\ Methods\end{tabular}} & DGER~\citep{zhao2020domain}       & ResNet18                  & 80.70 & 76.40 & 96.65 & 71.77 & 81.38   \\
                                                                                & DIRT-GAN~\citep{nguyen2021domain}   & ResNet18                  & 82.56 & 76.37 & 95.65 & 79.89 & 83.62   \\ \midrule
\multirow{5}{*}{\begin{tabular}[c]{@{}c@{}}Federated\\Learning\\ Methods \end{tabular}}           & FedAvg     & ResNet18                  & 55.83\textpm0.31 & 61.37\textpm0.66 & 77.87\textpm0.61 & 74.53\textpm0.18 & 67.40   \\
                                                                                & FedProx    & ResNet18                  & 56.84\textpm0.88       & 62.56\textpm0.87      & 78.33\textpm0.46      & 75.17\textpm0.61      & 68.23         \\
                                                                                & DRFA       & ResNet18                  & 56.59\textpm0.34 & 62.87\textpm0.22 & 78.63\textpm0.77 & 75.55\textpm0.42 & 68.41   \\
                                                                                & FedSR      & ResNet18                  & 57.56\textpm0.95     & 61.91\textpm0.35      & 78.42\textpm0.19      &  74.73\textpm0.27     & 68.16        \\
                                                                                & TFL (Ours) & ResNet18                  & \textbf{59.05}\textpm0.69 & \textbf{64.46}\textpm0.21 & \textbf{79.35}\textpm0.61 & \textbf{76.93}\textpm0.39 & \textbf{69.95}        \\ 
\bottomrule
\end{tabular}}
\label{tab:pacs}
\end{table*}

\begin{table*}[ht]
\vskip -0.1in
\centering
\caption{Best ROC-AUCs, corresponding communication round, and the total number of communicated parameters (donate as c-params. (M) in the table) on eICU. This dataset comprises EHRs collected from a diverse range of $72$ hospitals across the United States. We trained our model using data from $58$ hospitals located in the MIDWEST, NORTHEAST, and SOUTH regions. We evaluated the performance of the global model on an independent set of $14$ hospitals from the WEST. The reported numbers are from three independent runs. \textit{ Our approach achieves the best OOF performance with minimal communication rounds and the number of communicated parameters.}}
\vskip 0.1in
\resizebox{\textwidth}{!}{
\begin{tabular}{c|cclcclcclcclccl}
\toprule
Centralized Method   & \multicolumn{15}{c}{Federated Learning Methods}                                                                                                                                                                                                                                                   \\ \midrule
\multirow{2}{*}{ERM} & \multicolumn{3}{c|}{FedAvg}                              & \multicolumn{3}{c|}{FedProx}                             & \multicolumn{3}{c|}{DRFA}                                & \multicolumn{3}{c|}{FedSR}                               & \multicolumn{3}{c}{TFL (Ours)}                        \\ \cmidrule{2-16} 
                     & ROC-AUC           & round & \multicolumn{1}{l|}{c-params.} & ROC-AUC           & round & \multicolumn{1}{l|}{c-params.} & ROC-AUC           & round & \multicolumn{1}{l|}{c-params.} & ROC-AUC           & round & \multicolumn{1}{l|}{c-params.} & ROC-AUC          & round & c-params.                    \\ \midrule
67.04\textpm1.88    & 57.18\textpm0.03 & 6     & \multicolumn{1}{c|}{15.66M}  & 57.21\textpm0.01 & 6     & \multicolumn{1}{c|}{15.66M}  & 57.20\textpm0.09 & 2     & \multicolumn{1}{c|}{10.44M}  & 57.25\textpm0.03 & 8     & \multicolumn{1}{c|}{20.88M}  & \textbf{58.41}\textpm0.06 & 2     & \multicolumn{1}{c}{10.44M} \\ \bottomrule
\end{tabular}}
\label{tab:eicu}
\vspace{-10pt}
\end{table*}

\textbf{Datasets.} TFL is evaluated on our curated real-world datasets (\ding{172}eICU, \ding{173}FeTS, \ding{174}TPT-48) and standard benchmarks (\ding{175}CIFAR-10/-100, \ding{176}PACS), spanning a wide range of tasks including classification, regression, and segmentation. Evaluations covers both out-of-federation (datasets \ding{172}-\ding{174}, \ding{176}) and in-federation (datasets \ding{174}-\ding{175}) scenarios. Below are the descriptions of the real-world datasets. For detailed descriptions of all datasets, including \ding{175}CIFAR-10/-100 and \ding{176}PACS, please refer to Supplementary A.

\begin{enumerate}[leftmargin=13pt, noitemsep, topsep=0pt, label=\ding{\numexpr171+\arabic*\relax}]
\setlength\itemsep{2.5pt}
    \item \textbf{eICU}~\citep{pollard2018eicu} is a large-scale multi-center critical care database. It contains high granularity critical care data for over $200,000$ patients admitted to $208$ hospitals across the United States. We follow~\citep{huang2019patient} to pre-process the raw data. The resultant dataset comprises $19,000$ patients over $72$ hospitals. We further split the patients into four groups based on census regions and provided the protocol for OOF evaluation.
    

    
    \item \textbf{FeTS}~\citep{pati2022fets} is a multi-institutional medical imaging dataset. It comprises clinically acquired MRI scans of glioma. We use a subset of the original data, comprising $358$ subjects from $21$ distinct global institutions, and provide a protocol for OOF evaluation. The task is brain tumor segmentation and the evaluation metric is Dice Similarity Coefficient (DSC $\uparrow$). 
    
    \item  \textbf{TPT-48}~\citep{vose2014gridded} contains the monthly average temperature for the 48 contiguous states in the US from 2008 to 2019. The task is to predict the next six months’ temperature given the first six months’ data. We consider two tasks: (1) E(24) $\rightarrow$ W(24): Using the 24 eastern states as IF clients and the 24 western states as OOF clients; (2) N(24) $\rightarrow$ S(24): Using the 24 northern states as IF clients and the 24 southern states as OOF clients. The evaluation metric is Mean Squared Error (MSE $\downarrow$).
    
    
\end{enumerate}

\textbf{Baselines}. We compare with FedAvg~\citep{mcmahan2017communication} and FedProx~\citep{li2020federated}, the most referenced baselines in the literature. DRFA~\citep{deng2020distributionally} is the latest work that adopts the federated distributionally robust optimization framework. FedSR~\citep{nguyen2022fedsr} is the most recent work that tackles FL's generalization to unseen clients. We did not compare with FedDG~\citep{liu2021feddg} as it requires the sharing of data in the frequency space with each other. This can be viewed as a form of data leakage~\cite{bai2023benchmarking,nguyen2022fedsr}.  We provide implementation details on model architecture and hyperparameters in Supplementary D.

\subsection{Evaluation on OOF-resiliency}
\label{exp:LoT}

\begin{figure}[t]
\begin{center}
\includegraphics[width=0.98\linewidth]{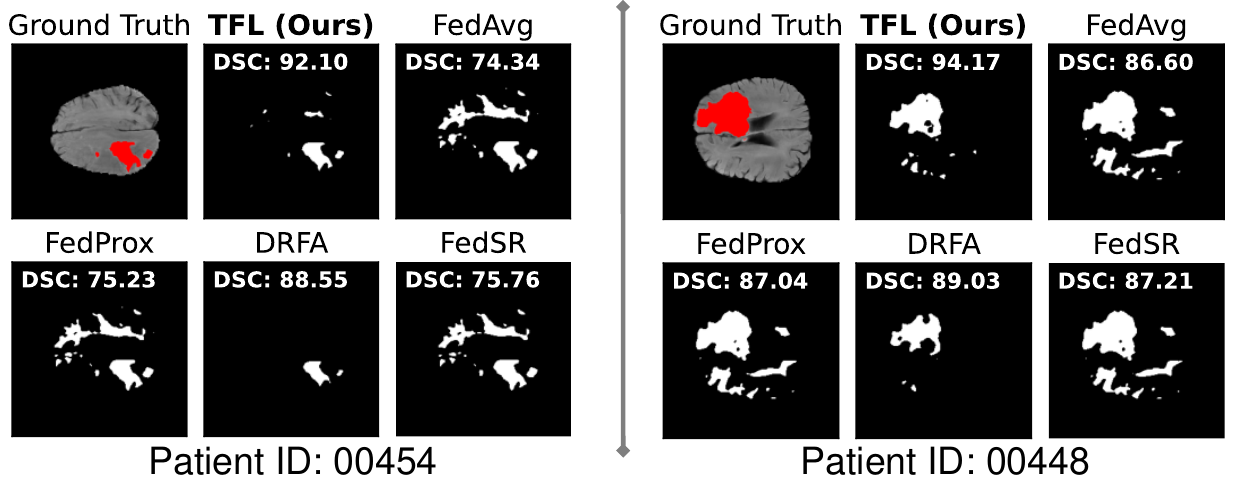}
\end{center}
\vskip -0.15in
\caption{Qualitative results on unseen patients from FeTS dataset. We also show the DSC ($\uparrow$) score. {\it Our method consistently demonstrates superior OOF robustness under diverse local demographics}. Additional visualizations can be found in Supplementary A.}
\vskip -0.2in
\label{fig:seg}
\end{figure}

\textbf{Takeaway 1: Learning on client topology improves OOF robustness}.
We evaluate our method on different datasets and summarize the results in Tables~\ref{tab:pacs},~\ref{tab:eicu}, and~\ref{tab:fets}. We make the following observation: our method improves the model's OOF-resiliency on both standard and real-world datasets. From Table~\ref{tab:pacs}, our method performs $2.25\%$ better than the federated robust optimization method (DRFA) and $2.63\%$ better than the federated domain generalization method (FedSR).
From Table~\ref{tab:eicu}, our method performs $2.11\%$ better than DRFA and $2.03\%$ better than FedSR.
Lastly, from Table~\ref{tab:fets}, our method performs $1.98\%$ better than DRFA and $3.00\%$ better than FedSR. Overall, our method shows consistently superior OOF robustness than state-of-the-art across all the evaluated datasets. We also visualize the segmentation results of FeTS in Figure~\ref{fig:seg}. 
Our method delivers high-quality segmentation results, suggesting increased reliability for real-world healthcare applications that have to contend with diverse local demographics.

\begin{figure*}[ht]
\begin{center}
\includegraphics[width=0.98\linewidth]{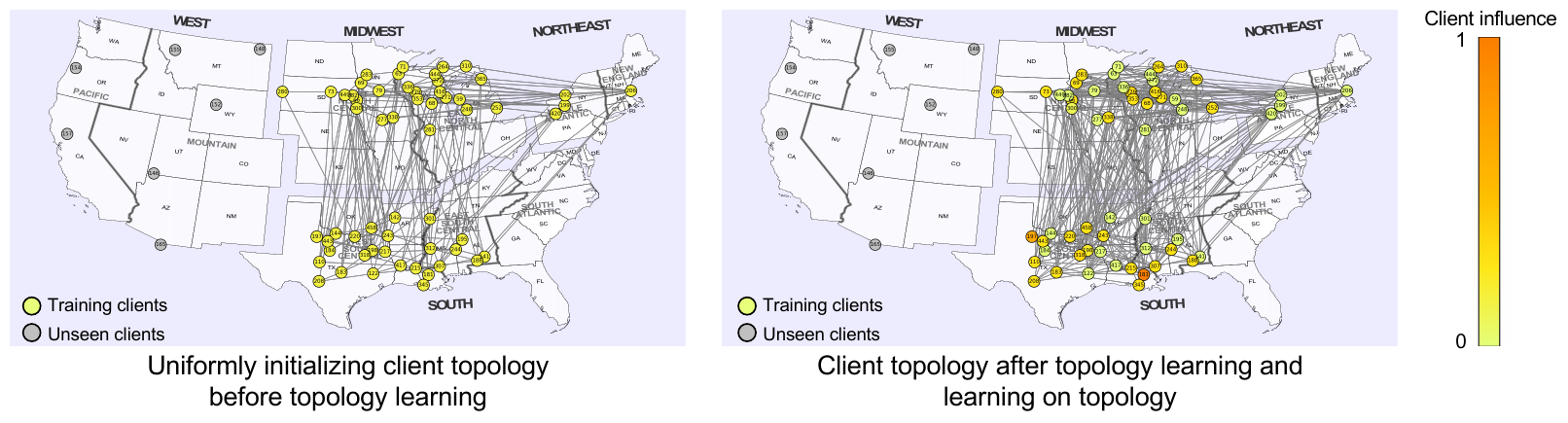}
\end{center}
\vspace{-10pt}
\caption{Visualization of client topology on the eICU dataset. Hospitals from MIDWEST, SOUTH, and NORTHEAST collaboratively train a global model, which is then applied to hospitals in the WEST region for OOF generalization. We observe that the learned client topology becomes denser, revealing underlying relationships that were previously unknown. Furthermore, the identified ``influential'' hospitals are more from the MIDWEST and SOUTH rather than NORTHEAST. This observation is rational, given the geographical proximity of these two regions to the target evaluation region, the WEST.}
\vspace{-10pt}
\label{fig:topology}
\end{figure*}

\textbf{Visualization of client topology}. We also visualize the learned client topology of the eICU dataset in Figure~\ref{fig:topology}. We observe that the learned client topology helps to identify the ``influential'' clients. The figure indicates that the majority of important clients are from the MIDWEST and SOUTH, with no influential clients from the NORTHEAST.

\begin{table}[t]
\vskip -0.1in
\centering
\caption{DSC ($\uparrow$) score on FeTS. This dataset contains tumor images from 21 institutions worldwide. We conduct training on 15 institutions and evaluate the model on the remaining 5. The reported numbers are from three independent runs. \textit{Our method delivers the best OOF robustness.}}
\vskip 0.1in
\resizebox{\linewidth}{!}{\begin{tabular}{c|ccccc}
\toprule
Centralized Method & \multicolumn{5}{c}{Federated Learning Methods}                                                                                                                     \\ \midrule
ERM                                                         & FedAvg                           & FedProx & DRFA                             & FedSR & TFL (Ours)                                                 \\ \midrule
                                             83.14\textpm {0.98}                & 71.45\textpm {0.05} &  71.15\textpm {0.04}        & 72.12\textpm {1.03} &   72.85\textpm {0.05}     & \textbf{74.29}\textpm {0.15} \\ \bottomrule
\end{tabular}}
\vspace{-5pt}
\label{tab:fets}
\end{table}

\subsection{Evaluation on Scalability}

\begin{figure}
    \includegraphics[trim=0cm 0cm 0cm 0.0cm,clip, width=0.45\textwidth]{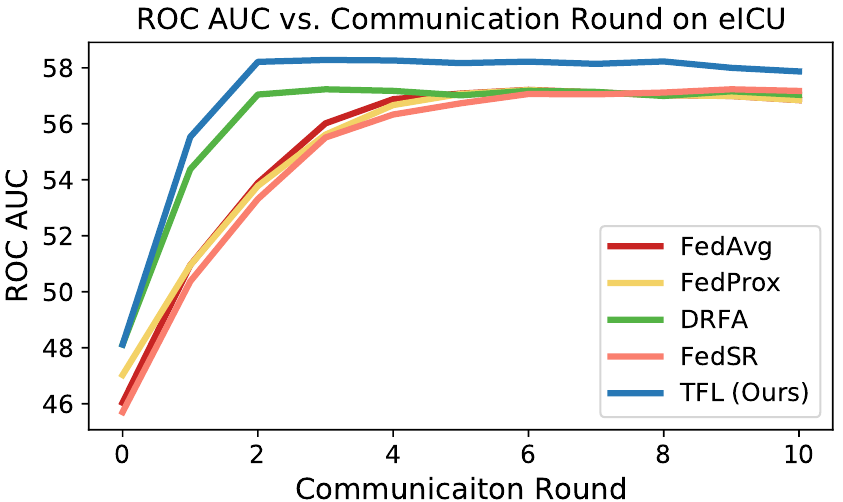}
    \caption{Visualization of ROC AUC vs. comm. round. \textit{Our method yields the best result with fewer communication rounds}.}
  \vspace{-15pt}
  \label{fig:commvsacc}
\end{figure}

\textbf{Takeaway 2: TFL is scalable for OOF generalization.}
We investigate whether TFL will significantly increase the communication or computation overhead, potentially affecting scalability in large-scale settings. We present ROC AUC, communication rounds, and transmitted parameters for the eICU dataset in Table~\ref{tab:eicu} and Figure~\ref{fig:commvsacc}. Our method achieves the highest ROC AUC score while requiring the fewest communication rounds and the total number of communicated parameters, thereby indicating its superior communication efficiency. Notably, DRFA also achieves its peak performance within the same number of communication rounds as our method. This is primarily attributable to the fact that both DRFA and our approach utilize a distributionally robust optimization framework. By minimizing the worst-case combination of local client risks, the model converges faster toward the optimum. Additionally, we also show the wall clock time versus OFF accuracy on PACS in Figure~\ref{fig:tradeoff}. Our method approximates the scalability of FedAvg and FedSR while achieving superior OOF accuracy.

\subsection{Evaluation on In-federation Performance}
\textbf{Takeaway 3: Prioritizing influential clients also benefit in-federation (IF) performance.}
We have shown that TFL can significantly boost the model's OOF robustness. This naturally leads us to examine its IF performance. We are interested in understanding if TFL can also improve the IF performance. To this end, we conducted empirical evaluations using the TPT-48 and CIFAR-10/-100. The TPT-48 dataset, offering additional hold-out data from IF clients, is particularly well-suited for evaluating IF performance. CIFAR-10 and -100 are widely recognized benchmarks for IF evaluation. The results, shown in Tables~\ref{table:tpt48_r} and~\ref{table:cifar_r}, reveal that our method indeed enhances the IF performance. For example, TFL outperforms DRFA by $2.2\%$ on TPT-48 and $2.5\%$ on CIFAR-10. These findings underscore the value of focusing on influential clients when building robust models with good IF and OOF performance.

\begin{table}[t]
\vskip -0.1in
    \centering
    \caption{Results on TPT-48. We report results for both IF and OOF evaluations. The numbers are the mean of three independent runs. \textit{Our method improves both the OOF and IF performance.}}
    \vskip 0.1in
    \resizebox{\linewidth}{!}{\begin{tabular}{cccccc}
\toprule
\multicolumn{1}{c|}{Method} & \multicolumn{1}{c|}{Centralized} & \multicolumn{1}{c}{FedAvg} & \multicolumn{1}{c}{FedProx} & \multicolumn{1}{c}{DRFA}   & TFL (Ours)    \\ \midrule
\multicolumn{6}{c}{\textit{Out-of-federation} Evaluation (MSE $\downarrow$)}                                                                                                            \\ \midrule
\multicolumn{1}{c|}{E(24) $\rightarrow$ W(24)} & \multicolumn{1}{c|}{0.3998}      & \multicolumn{1}{c}{0.6264} & \multicolumn{1}{c}{0.6312}  & \multicolumn{1}{c}{0.5451} & \textbf{0.4978} \\ 
\multicolumn{1}{c|}{N(24) $\rightarrow$ S(24)} & \multicolumn{1}{c|}{1.4489}      & \multicolumn{1}{c}{2.0172} & \multicolumn{1}{c}{1.9729}  & \multicolumn{1}{c}{1.8972} & \textbf{1.7432} \\ \midrule \midrule
\multicolumn{6}{c}{\textit{In-federation} Evaluation (MSE $\downarrow$)}                                                                                                                \\ \midrule 
\multicolumn{1}{c|}{E(24) $\rightarrow$ E(24)} & \multicolumn{1}{c|}{0.1034}      & \multicolumn{1}{c}{0.2278} & \multicolumn{1}{c}{0.2163}  & \multicolumn{1}{c}{0.1554} & \textbf{0.1523} \\
\multicolumn{1}{c|}{N(24) $\rightarrow$ N(24)} & \multicolumn{1}{c|}{0.1329}      & \multicolumn{1}{c}{0.1550} & \multicolumn{1}{c}{0.1523}  & \multicolumn{1}{c}{0.1437} & \textbf{0.1405} \\ \bottomrule
\end{tabular}}
\label{table:tpt48_r}
\end{table}

\begin{table}[t]
\vskip -0.1in
    \centering
    \caption{In-federation accuracy on CIFAR-10/-100, where a Dirichlet distribution~\citep{hsu2019measuring} with parameter Dir(0.1) is used to partition the datasets into heterogeneous clients at different scales. The reported numbers are the mean of three independent runs. \textit{Our method yields the best accuracy at various scales.}}
    \vskip 0.1in
    \resizebox{\linewidth}{!}{\begin{tabular}{c|c|cccc}
\toprule
Dataset                    & $\#$ of clients & FedAvg & FedProx & DRFA  & TFL (Ours) \\ \midrule
\multirow{2}{*}{CIFAR-10}  & 25              & 68.07  & 68.15   & 70.51 & \textbf{72.24}      \\ 
                           & 50              & 64.48  & 64.33   & 63.19 & \textbf{65.59}      \\ \midrule\midrule
\multirow{2}{*}{CIFAR-100} & 25              & 37.90  & 37.76   & 38.04 & \textbf{38.85}      \\ 
                           & 50              & 37.33  & 37.28   & 37.54 & \textbf{38.08}      \\ \bottomrule
\end{tabular}}
\label{table:cifar_r}
\vskip -0.1in
\end{table}

\subsection{Evaluation on Effectiveness of Client Clustering}
\textbf{Takeaway 4: Client clustering reduces the computation costs of client topology learning.}
We conducted experiments on eICU to validate whether client clustering effectively reduces computation costs. The eICU dataset was chosen due to its larger scale (including 72 clients) compared to all other evaluated datasets. Specifically, during client topology learning, we employed KMeans~\cite{lloyd1982least} to group training clients into several clusters (\textit{e.g.,} 10 clusters) based on their model weights. The client topology is then learned at the cluster level. As shown in Table~\ref{tab:c_clustering}, our clustering approach significantly reduces computation costs while slightly decreasing the OOF accuracy.

\begin{table}[t]
    \centering
    \caption{Comparison of wall-clock time on eICU. \textit{Our client clustering approach (with 10 clusters) significantly decreases computation costs by $69\%$, while only marginally impacting OOF performance, evidenced by a slight reduction of $0.77\%$}.}
    \vskip 0.1in
\resizebox{0.9\linewidth}{!}{\begin{tabular}{l|c|c}
\toprule
                  & \multicolumn{1}{l|}{ROC-AUC} & \multicolumn{1}{l}{Wall-clock time (s)} \\ \midrule
FedAvg            & 57.18\textpm0.03               & 120.15                               \\ 
TFL               & 58.41\textpm0.06               & 437.61                               \\ 
TFL w/ Clustering & 57.96\textpm0.18               & 133.08                               \\ \bottomrule
\end{tabular}}
\label{tab:c_clustering}
\vskip -0.15in
\end{table}

\subsection{Ablation Study}
\label{exp:ablation}

\textbf{Effects of partial model approximation.} To handle high-dimensional models,  we can leverage partial model parameters to compute the similarity scores for better computation efficiency. Here, we report the OOF performance when using full and partial model parameters. Experiments are conducted on PACS and results are shown in Table~\ref{tab:approx}.

\textbf{Effects of graph sparsity.} To avoid learning a fully connected graph with trivial edges, we add a sparsity contain to client topology learning. In implementation, we adopt $\epsilon$-graph to ensure Equation~\ref{eqn:ctl} is solvable. The threshold value $\epsilon$ controls the graph sparsity. Here, we investigate how it will affect the TFL. We report the results on PACS in Table~\ref{fig:ablation}. We observe that $\epsilon=0.4$ yields the best performance.

\textbf{Effects of topology update frequency.} The frequency of updating client topology during training affects the total training time ($f=5$ denotes updating the topology every 5 rounds). More frequent updates result in extended training times. We investigated the impact of client topology updating frequency on the performance of the model on PACS. Our results in Table~\ref{fig:ablation} indicate that model performance remains relatively stable across various updating frequencies. \textbf{We provide more ablation studies in Supplementary C}.

%% file: sec_submit/sec5_releated.tex
\section{Related Work}


\textbf{FL generalization to unseen clients.} There are recent attempts to address generalization to unseen clients in FL. FedDG~\cite{liu2021feddg} is proposed to share the amplitude spectrum of images among local clients to augment the local data distributions. FedADG~\cite{zhang2021federated} adopts the federated adversarial training to measure and align the local client distributions to a reference distribution. FedSR~\cite{nguyen2022fedsr} proposes regularizing latent representation's $\ell_{2}$ norm and class conditional information to enhance the OOF performance. However, existing methods often ignore scalability issues, yielding inferior performance in large-scale distributed setting~\cite{bai2023benchmarking}. We introduce an approach that employs client topology to achieve good OOF-resilency in a scalable manner.

\textbf{FL with Graphs.} One line of research is to combine FL with GNNs, leading to federated GNNs~\cite{xie2021federated,wu2021fedgnn}. These federated GNNs primarily engage in learning from distributed graph data that already has inherent structural information. In contrast, our primary objective is to obtain a distribution graph that captures the relationships between distributed clients. Another research line explores the use of graphs to enhance personalization in FL~\cite{ye2023personalized,zantedeschi2019fully}. For example, pFedGraph~\cite{ye2023personalized} utilizes a collaboration graph to tailor the global model more effectively to individual clients. These types of research primarily deal with in-federation clients. Different from these methods, our approach aims to utilize a client topology to improve the FL model's generalizability to clients outside of the federation.


\textbf{Graph topology learning.} The problem of graph topology learning has been studied in graph signal processing~\cite{stankovic2020data} and graph neural networks~\cite{chen2020iterative,chen2020Reinforcement}. Traditional topology learning methods often require centralizing the data, raising privacy concerns. Yet, how to estimate the graph topology with privacy regulations has been less investigated. We explore simple methods to infer the graph topology using model weights. \textbf{An extended related work section is provided in Supplementary B.}

\begin{table}[t]
\centering
\caption{TFL's OOF accuracy on PACS using full and partial model parameters. The results show that \textit{using partial model parameters achieves performance on par with full parameters}.}
\vskip 0.1in
\resizebox{0.95\linewidth}{!}{\begin{tabular}{cccccc}
\toprule
        & A     & C     & P     & S     & Avg. \\ \midrule
Full    & 59.05\textpm{\scriptsize 0.69} & 64.46\textpm{\scriptsize 0.21} & 79.35\textpm{\scriptsize 0.61} & 76.93\textpm{\scriptsize 0.39} & 69.95   \\ \midrule
Partial & 58.96\textpm{\scriptsize 0.85}      &  64.57\textpm{\scriptsize 0.98}     & 78.94\textpm{\scriptsize 0.65}      & 76.61\textpm{\scriptsize 0.13}      & 69.77         \\ \bottomrule
\end{tabular}}
\label{tab:approx}
\vspace{-10pt}
\end{table}

\begin{table}[t]
    \centering
        \caption{Ablation study on effects of graph sparsity and client topology update frequency on PACS.}
        \vskip 0.1in
    \resizebox{0.75\linewidth}{!}{\begin{tabular}{ccccc}
\toprule
\multicolumn{5}{c}{Graph Sparsity ($\epsilon$), Accuracy $\uparrow$}       \\ \midrule
$\epsilon$ = 0.45        & $\epsilon$ = 0.4       & $\epsilon$ = 0.38         & $\epsilon$ = 0.35         & $\epsilon$ = 0.3           \\ \midrule
58.61      & \textbf{59.05}      & 58.83        & 58.11        & 57.73          \\ \midrule \midrule
\multicolumn{5}{c}{Topology update frequency ($f$), Accuracy $\uparrow$}       \\ \midrule
$f$ = 50        & $f$ = 30       & $f$ = 20         & $f$ = 10         & $f$ = 5           \\ \midrule
\textbf{59.19}      & 58.91      & 58.27        & 58.80        & 58.28          \\ \bottomrule
\end{tabular}}
\label{fig:ablation}
\vspace{-15pt}
\end{table}

%% file: sec_submit/sec6_conclusion.tex
\section{Conclusion}

Federated Learning faces significant out-of-federation (OOF) generalization challenges that can severely impair model performance. We propose to improve OOF robustness by leveraging client relationships, leading to \textit{Topology-aware Federated Learning} (TFL). TFL contains two key modules: i) Inferring a topology that describes client relationships with model similarity and ii) Leveraging the learned topology to build a robust model. Extensive experiments on real-world and benchmark datasets show that TFL demonstrates superior OOF-resiliency with scalability.

%% file: sec_submit/sec7_impack.tex
\section*{Acknowledgment} This work is supported by the National Science Foundation through the Faculty Early Career Development Program (NSF CAREER) Award, the Department of Defense under the Defense Established Program to Stimulate Competitive Research (DoD DEPSCoR) Award, and the University of Delaware Artificial Intelligence Center of Excellence (AICoE) Seed Grant.

\section*{Impact Statement}

\textbf{Societal impact.} Our research has the potential to impact various sectors, notably automated vehicles (AVs), healthcare, and finance, through the enhancement of Federated Learning (FL) models. By improving the robustness of FL models for unseen clients, we anticipate a range of societal benefits. In healthcare, this could manifest as more accurate medical diagnostics, leading to better patient outcomes. In the realm of AVs, enhanced FL models could result in safer navigation and operation, particularly under challenging weather or road conditions. In finance, these advancements could foster better inter-company collaboration, leading to more efficient and secure financial services. 

\textbf{Ethical considerations.} While we initially did not identify the potential for our research to lead to unintended biases or discrimination, we recognize the complexity and evolving nature of ethical challenges in Machine Learning. It is important to continuously evaluate our models and methodologies for any inadvertent biases, especially as they are applied in diverse, real-world scenarios. We commit to rigorously testing our models to identify and mitigate any such biases, ensuring that our research contributes positively and equitably across all sections of society.

%% file: sec8_supps.tex
\appendix
\onecolumn


\section{Additional Experimental Results \& Detailed Dataset Description}
\subsection{Evaluating OOF-resiliency on OfficeHome}
We conduct additional experiments on \textbf{OfficeHome}~\citep{venkateswara2017deep} dataset. It contains $15,588$ images from four domains: art, clipart, product, and real world. The task is a $65$-class classification problem. Like PACS's experimental setup, we evenly split each domain into $5$ subsets, yielding $20$ subsets, and treat each subset as a client. We followed the common \textit{leave-one-domain-out} experiment, where $3$ domains are used ($15$ clients) for training and $1$ domain ($5$ clients) for testing. We use ResNet50~\citep{he2016deep} as our model and train the model for $100$ communication rounds. Each local client optimized the model using stochastic gradient descent (SGD) with a learning rate of $0.01$, a momentum of $0.9$, weight decay of $5e^{-4}$, and a batch size of $64$. The model is evaluated using classification accuracy.


\begin{table}[th]
\centering
\caption{Accuracy on the \textbf{OfficeHome} dataset. We conduct experiments using a \textit{leave-one-domain-out} approach, meaning each domain serves as the evaluation domain in turn. Existing methods typically consider each domain as an individual client~\citep{liu2021feddg,nguyen2022fedsr}. However, in order to simulate a large-scale distributed setting, we took a different approach by further dividing each domain into $5$ subsets and treating each subset as a separate client. This increased the total number of clients to $20$. \textit{Our method outperformed others across all experimental settings}, demonstrating superior results. } 
\vskip 0.1in
\resizebox{0.85\linewidth}{!}{
\begin{tabular}{cccccccc}
\toprule
\multicolumn{2}{c}{\multirow{2}{*}{Models}}                                                  & \multirow{2}{*}{Backbone} & \multicolumn{5}{c}{OfficeHome}                \\ \cmidrule{4-8} 
\multicolumn{2}{c}{}                                                                         &                           & A     & C     & P     & R     & Average \\ \midrule
\multirow{2}{*}{\begin{tabular}[c]{@{}c@{}}Centralized \\ Methods\end{tabular}} & Mixup~\citep{xu2020adversarial}       & ResNet50                  & 64.7 & 54.7 & 77.3 & 79.2 & 69.0   \\
                                                                                & CORAL~\citep{sun2016coral}   & ResNet50                  & 64.4 & 55.3 & 76.7 & 77.9 & 68.6   \\ \midrule
\multirow{5}{*}{\begin{tabular}[c]{@{}c@{}}Federated\\Learning\\ Methods\end{tabular}}           & FedAvg     & ResNet50                  & 24.10 & 23.16 & 40.19 & 43.47 & 32.73   \\
                                                                                & FedProx    & ResNet50                  &  23.16     & 23.47       &  41.08     &  42.66     &   32.59      \\
                                                                                & DRFA       & ResNet50                  & 25.29 & 23.98 & 41.23 & 42.35 & 
                                                                                33.21  \\
                                                                                & FedSR      & ResNet50                  & 23.51      & 22.93     &  39.30     &    41.48   &   
                                                                                31.81      \\
                                                                                & TFL (Ours) & ResNet50                  & \textbf{26.37} & \textbf{24.47} & \textbf{43.96} & \textbf{44.74} & \textbf{34.89}        \\ \bottomrule
\end{tabular}}
\end{table}

\subsection{Data Pre-processing on eICU}
We follow~\citep{huang2019patient} to predict patient mortality using drug features. These features pertain to the medications administered to patients during the initial 48 hours of their ICU stay. We've extracted pertinent patient and corresponding drug feature data from two primary sources: the 'medication.csv' and 'patient.csv' files. Our final dataset is a table with the dimension of $19000 \times 1411$. Each row in this matrix symbolizes a unique patient, while each column corresponds to a distinct medication.

\subsection{More Results on eICU}
\begin{figure*}[t]
\begin{center}
\includegraphics[width=0.9\linewidth]{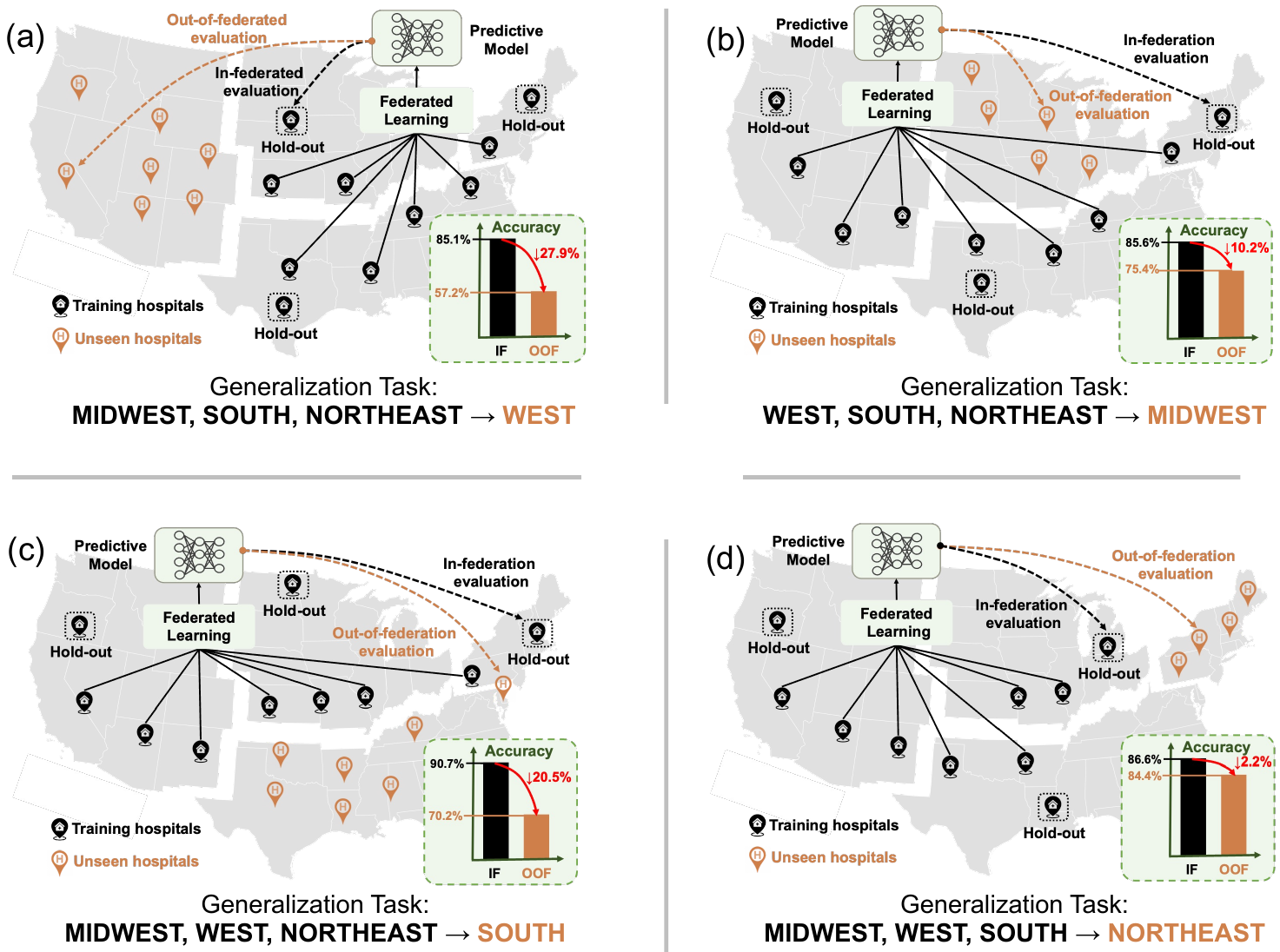}
\end{center}
\caption{Additional results on eICU (IF \textit{vs.} OOF performance). eICU's 72 hospitals are distributed across the United States. Specifically, there are $14$ hospitals in the WEST, $28$ hospitals in the MIDWEST, $26$ hospitals in the SOUTH, and $4$ hospitals in the NORTHEAST. We employ a \textit{leave-one-region-out} approach, designating one geographic region as the OOF region while the remaining as IF regions. We observe a considerable gap between IF and OOF performance, indicating that {\it current FL methods are not robust against OOF data}.
}
\label{fig:moregaps}
\end{figure*}

We conduct more experiments on the eICU dataset to evaluate the gap between in-federation (IF) and out-of-federation (OOF) and visualize the results in Figure~\ref{fig:moregaps}. \textit{We observe that existing FL methods are not robust against OOF data.} 

\subsection{Detailed Dataset Description}

\begin{figure*}[ht]
\begin{center}
\includegraphics[width=0.9\linewidth]{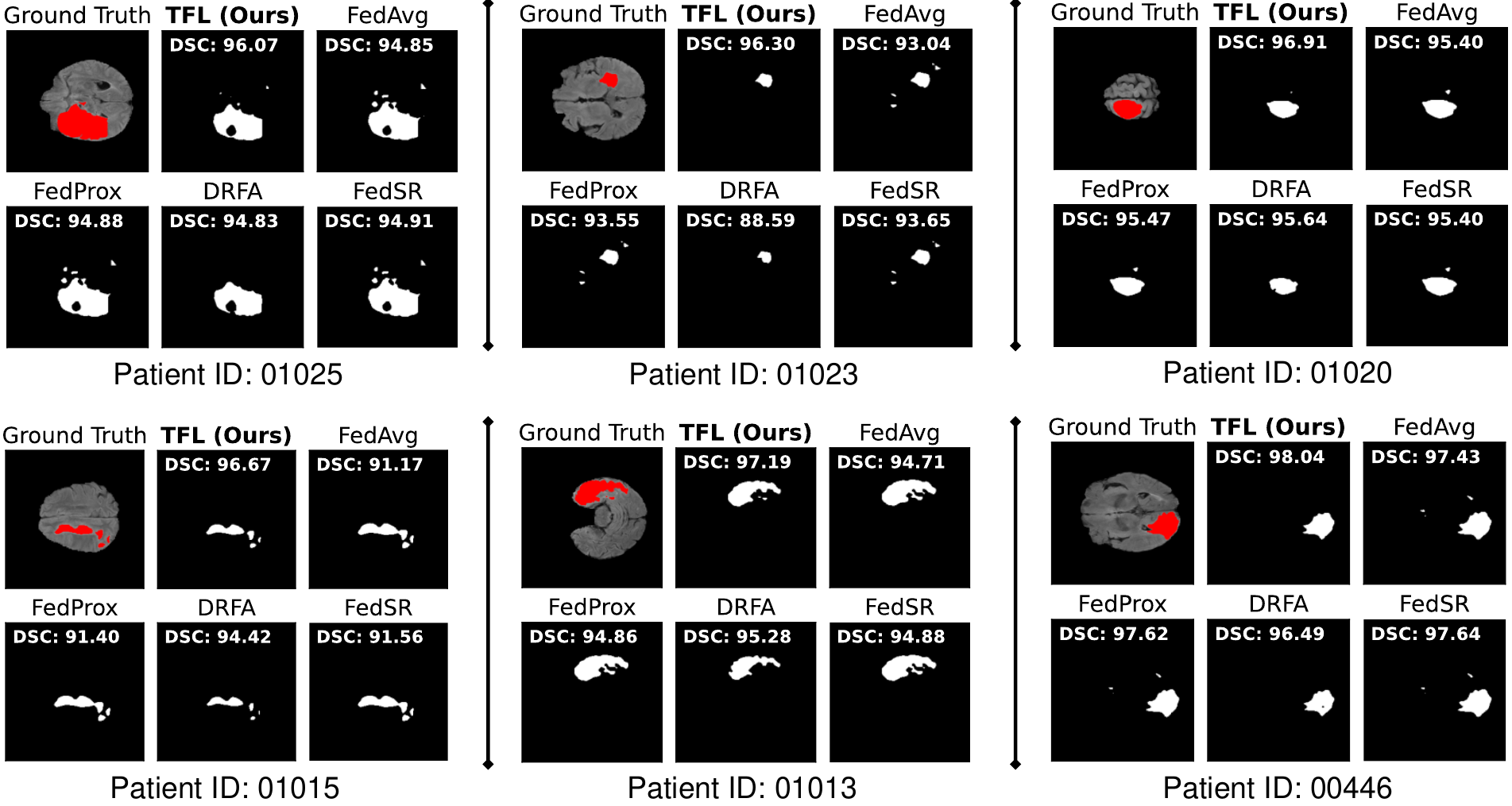}
\end{center}
\caption{Additional qualitative results comparison on unseen patients of the FeTS dataset. We show both the tumor segmentation and DSC ($\uparrow$) score.
{\it Our method demonstrates consistent superior OOF-resiliency across a range of local demographics.}
}
\vspace{-10pt}
\label{fig:seg_additional}
\end{figure*}

\vspace{-3pt}
\begin{enumerate}[leftmargin=13pt, noitemsep, topsep=0pt, label=\ding{\numexpr171+\arabic*\relax}]
\setlength\itemsep{2.5pt}
    \item \textbf{eICU} (eICU Collaborative Research Database)~\citep{pollard2018eicu} is a large-scale multi-center critical care database. It contains high granularity critical care data for over $200,000$ patients admitted to $208$ hospitals across the United States. Each hospital is considered an individual client. We follow~\citep{huang2019patient} to prepare the data. The final dataset includes data from approximately $19000$ patients over $72$ hospitals. 
    The generalization task could be deploying models trained on hospitals from the SOUTH region to those in the WEST. The evaluation metric for patient mortality prediction is ROC-AUC.
    
    \item \textbf{FeTS} (Federated Tumor Segmentation Dataset)~\citep{pati2022fets} is a multi-institutional medical imaging dataset. It comprises clinically acquired multi-institutional MRI scans of glioma. A subset of the original data was used, comprising $358$ subjects from $21$ distinct global institutions. The associated task is to identify and delineate brain tumor boundaries. Each institution is a client. Evaluation metric is Dice Similarity Coefficient (DSC $\uparrow$). See more visualizations in Figure~\ref{fig:seg_additional}.
    
    \item  \textbf{TPT-48}~\citep{vose2014gridded} contains the monthly average temperature for the 48 contiguous states in the US from 2008 to 2019. The task is to predict the next six months’ temperature given the first six months’ temperature. For TPT-48, we consider two generalization tasks: (1) E(24) $\rightarrow$ W(24): Using the 24 eastern states as IF clients and the 24 western states as OOF clients; (2) N(24) $\rightarrow$ S(24): Using the 24 northern states as IF clients and the 24 southern states as OOF clients. The evaluation metric is Mean Squared Error (MSE $\downarrow$).
    
    \item \textbf{CIFAR-10/-100}~\citep{krizhevsky2009learning} are the most commonly used benchmarks in FL literature. In these datasets, we use Dirichlet distribution~\citep{hsu2019measuring} to partition the dataset into the heterogeneous setting with $25$ and $50$ clients.
    
    \item \textbf{PACS}~\citep{li2017pacs} contains $9,991$ images from four domains: art painting, cartoon, photo, and sketch. The task is seven-class classification. For PACS, we evenly split each domain into $5$ subsets, yielding $20$ subsets, and we treat each subset as a client. We followed the common ``leave-one-domain-out" experiment, where $3$ domains are used ($15$ clients) for training and $1$ domain ($5$ clients) for testing. Model performance is evaluated by classification accuracy.
\end{enumerate}

\vspace{-5pt}

\section{Extended Related Work}

\textbf{Federated learning.} Federated learning~\cite{li2020federatedspm,kairouz2021advances,mcmahan2017communication} has emerged as a powerful tool to protect data privacy in the distributed setting. It allows multiple clients/devices to collaborate in training a predictive model without sharing their local data. 
Despite the success, current FL methods are vulnerable to heterogeneous data (non-IID data)~\cite{smith2017federated,sattler2019robust,li2023data,tan2022fedproto}, a common issue in real-world FL. Data heterogeneity posits significant challenges to FL, such as the severe convergence issue~\cite{li2020on} and poor generalization ability to new clients~\cite{sattler2019robust}. To improve the model's robustness against data heterogeneity, \texttt{FedProx}~\cite{li2020federated} add a proximal term to restrict the local model updating, avoiding biased models toward local data distribution. \texttt{SCAFFOLD}~\cite{karimireddy2020scaffold} introduces a control variate to rectify the local update. \texttt{FedAlign}~\cite{mendieta2022local} improve the heterogeneous robustness by training local models with better generalization ability. However, most FL methods focus on the model's in-federation performance. Orthogonal to existing work, we propose leveraging client topology to improve the model's OOF generalization capability.


\textbf{FL generalization to unseen clients.} There are recent attempts to address generalization to unseen clients in FL. FedDG~\cite{liu2021feddg} is proposed to solve domain generalization in medical image classification. The key idea is to share the amplitude spectrum of images among local clients to augment the local data distributions. FedADG~\cite{zhang2021federated} adopts the federated adversarial training to measure and align the local client distributions to a reference distribution. FedGMA~\cite{tenison2022gradient} proposes gradient masking averaging to prioritize gradients aligned with the overall domain direction across clients. FedSR~\cite{nguyen2022fedsr} proposes regularizing latent representation's $\ell_{2}$ norm and class conditional information to enhance the OOF performance. However, existing methods often ignore scalability issues, yielding inferior performance in large-scale distributed setting~\cite{bai2023benchmarking}. We introduce an approach that employs client topology to achieve good OOF-resiliency in a scalable manner.

\textbf{Graph topology learning.} The problem of graph topology learning has been studied in different fields. In graph signal processing~\cite{mateos2019connecting,dong2019learning,stankovic2020data}, existing work explores various ways to learn the graph structure from data with structural regularization ( \textit{e.g.,} sparsity, smoothness, and community preservation~\cite{zhu2021deep}). In Graph Neural Networks (GNNs)~\cite{wu2020comprehensive,welling2016semi}, researchers have explored scenarios where the initial graph structure is unavailable, wherein a graph has to be estimated from objectives~\cite{li2018adaptive,norcliffe2018learning} or words~\cite{chen2019graphflow,chen2020Reinforcement}. The existing graph topology learning methods often require centralizing the data, making it inapplicable in federated learning.
However, how to estimate the graph topology with privacy regulations has been less investigated. We explore simple methods to infer the graph topology using model weights. 

\textbf{FL with graphs.} One line of research is to combine FL with GNNs, leading to federated GNNs~\cite{xie2021federated,wu2021fedgnn}. These federated GNNs primarily engage in learning from distributed graph data that already has inherent structural information. In contrast, our primary objective is to obtain a distribution graph that captures the relationships between distributed clients. Another research line explores the use of graphs to enhance personalization in FL~\cite{ye2023personalized,zantedeschi2019fully}. For example, pFedGraph~\cite{ye2023personalized} utilizes a collaboration graph to tailor the global model more effectively to individual clients. These types of research primarily deal with in-federation clients. Our approach, however, aims to utilize client topology to significantly improve generalization to clients outside of the federation.

\section{Additional Ablation Study}

\setlength{\columnsep}{5pt}%
\begin{wraptable}{r}{0.55\textwidth}
\vspace{-20pt}
\centering
\caption{Ablation study evaluating the efficacy of hyperparameter tuning, centrality, and similarity metric.}
\begin{center}
\resizebox{0.9\linewidth}{!}{\begin{tabular}{ccccc}
\toprule
\multicolumn{5}{c}{Effectiveness of $q$, ROC AUC $\uparrow$}                              \\ \midrule
$q$ =1.0 & $q$ = $1e^{-1}$ & $q$ = $1e^{-2}$ &$q$ = $1e^{-3}$ & $q$ = $1e^{-4}$ \\ \midrule 
57.91     & \textbf{58.31 }     & 57.43        & 57.36        & 57.29          \\ \midrule \midrule
\multicolumn{5}{c}{Effectiveness of centrality, ROC AUC $\uparrow$}                \\ \midrule
Betweenness & Degree     & Closeness     & Eigenvector  & Current flow   \\ \midrule
\textbf{58.28}      & 57.69      & 57.86        & 57.57        & 57.83          \\ \midrule
\multicolumn{5}{c}{Effectiveness of similarity measure, Accuracy $\uparrow$}       \\ \midrule \midrule
             & $\ell_1$    & $\ell_2$    & dot produt   & cosin   \\ \midrule
OOF Accuracy & 58.11 & 58.26 & 59.14 & 58.52 \\ \bottomrule
\end{tabular}}
\end{center}
\label{fig:ablation1}
\end{wraptable}\leavevmode
\textbf{Hyperparameter $q$.} We investigate the impact of hyperparameter $\eta$ on eICU. Our findings demonstrate that setting $q = 0.1$ yields the best results.
\textbf{Centrality.} We employed betweenness centrality to derive the topological prior. However, it is worth noting that other types of centrality, such as degree~\citep{freeman2002centrality} and closeness~\citep{bavelas1950communication}, could also be utilized. We conducted experiments on eICU to verify the impact of different centrality measures on TFL. Our findings indicate that betweenness centrality produces the best result. \textbf{Similarity metrics.} We investigate how the model performs under different similarity metrics on PACS. We found that the dot product-based metric produces the best results.

\section{Implementation Details}

\textbf{Experiment settings and evaluation metrics.} For the \textbf{PACS dataset}, we evenly split each domain into $5$ subsets, yielding $20$ subsets, and we treat each subset as a client. We followed the common ``leave-one-domain-out" experiment, where $3$ domains are used ($15$ clients) for training and $1$ domain ($5$ clients) for testing. We evaluated the model's performance using classification accuracy.
We use ResNet18~\citep{he2016deep} as our model and train the model for $100$ communication rounds. Each local client optimized the model using stochastic gradient descent (SGD) with a learning rate of $0.01$, momentum of $0.9$, weight decay of $5e^{-4}$, and a batch size of $8$. 
For \textbf{CIFAR-10/100}, 
we adopt the same model architecture as {FedAvg}~\citep{mcmahan2017communication}. The model has 2 convolution layers with $32$, $64$ $5 \times 5$ kernels, and $2$ fully connected layers with $512$ hidden units. 
we use Dirichlet distribution~\citep{hsu2019measuring} to partition the dataset into the heterogeneous setting with $25$ and $50$ clients.
For \textbf{eICU}, we treat each hospital as a client. We use a network of three fully connected layers. This architecture is similar to~\citep{huang2019patient,ma2021smil,sheikhalishahi2020benchmarking}. 
We train our model for $30$ communication rounds, using a batch size of $64$ and a learning rate of $0.01$, and report the performance on unseen hospitals. Within each communication round, clients perform $5$ epochs (E = $5$) of local optimization using SGD. 
The evaluation metric employed was the ROC-AUC, a common practice in eICU~\citep{huang2019patient}. For \textbf{FeTS}, we treat each institution as a client. 
We adopt the widely used U-Net~\citep{ronneberger2015u} model. We train our model for $20$ communication rounds, using a learning rate of $0.01$ and a batch size of 64. We conduct training with 16 intuitions and report results on 5 unseen institutions. Each institution performs $2$ epochs of local optimization (E = $2$) using SGD. 
The evaluation metric is Dice Similarity Coefficient (DSC $\uparrow$). 
 For \textbf{TPT-48}, we consider two generalization tasks: (1) E(24) $\rightarrow$ W(24): Using the 24 eastern states as IF clients and the 24 western states as OOF clients; (2) N(24) $\rightarrow$ S(24): Using the 24 northern states as IF clients and the 24 southern states as OOF clients. 
 We use a model similar to ~\citep{xu2022graphrelational}, which has $8$ fully connected layers with $512$ hidden units. We use SGD optimizer with a fixed momentum of 0.9. 
 The evaluation metric is Mean Squared Error (MSE $\downarrow$).  Algorithm~\ref{alg:tafl} shows the overall algorithm of TFL. In implementation, we used dot product as the metric to measure client similarity. All experiments are conducted using a server with 8$\times$NVIDIA A6000 GPUs.
\begin{algorithm}[tb]
   \caption{Topology-aware Federated Learning}
   \label{alg:tafl}
\begin{algorithmic}
   \STATE {\bfseries Input:} $K$ clients; learning rate $\eta_{\theta}$ and $\eta_{\lambda}$; Communication round $T$; initial model $\theta^{(0)}$; initial $\lambda^{(0)}$.
   \REPEAT
   \FOR{each communication round $t=1, \cdots T$}
   \STATE server \textbf{samples} $m$ clients according to $\bm{\lambda^{(t)}}$
   \FOR{each client $i=1, \cdots m$ in parallel}
   \STATE $\theta^{t+1}_{i} = \theta^t_{i}  - \eta_{{\theta}^t_i} \nabla_{{\theta}^t_i} F({{\theta}^t_i})$
   \STATE client $i$ send $\theta^{t+1}_{i}$ back to the server
   \ENDFOR
   \STATE server \textbf{computes} $\bm{\theta}^{t+1} =\sum^{m}_{i=1} \theta^{t+1}_{i}$
   \IF{ $t \% f == 0$}
   \STATE server updating graph $\mathcal{G}$ via Equation 4
   \ENDIF
   \STATE server calculating topological prior $p$ from $\mathcal{G}$
   \STATE server calculating $\nabla_{\lambda^{(t)}} F(\bm{\theta}^{(t+1)}, \bm{\lambda^{(t)}})$ via Equation~\ref{eqn:tafl_un}
   \STATE server updates $\bm{\lambda}^{t+1} = \mathcal{P}_{\Delta_K} (\bm{\lambda}^t  + \eta_{\lambda}^t \nabla_{\lambda^{(t)}} F(\bm{\theta}^{(t+1)}, \bm{\lambda^{(t)}}))$
   \ENDFOR
   \UNTIL{convergence} 
\end{algorithmic}
\end{algorithm}

\section{Discussion of Limitations}

In this section, we discuss the limitations of TFL and the potential solutions.

\textbf{Concerns on privacy leakage.} Client topology learning may raise concerns about (unintentional) privacy leakage. However, we argue that any such leakage would be a general issue for FL methods rather than a unique concern for our approach. In comparison to standard FL, our method does not require additional information to construct the client topology, thus providing no worse privacy guarantees than well-established methods like FedAvg~\citep{mcmahan2017communication} and FedProx~\citep{li2020federated}. Nonetheless, FL may still be vulnerable to attacks that aim to extract sensitive information~\citep{bhowmick2018protection,melis2019exploiting}. In future work, we plan to explore methods for mitigating (unintentional) privacy leakage.

\textbf{Concerns on high-dimensional node embedding.} As outlined in Section 3.1, we harness model weights as node embeddings. Nevertheless, incorporating large-scale models, such as Transformers~\citep{vaswani2017attention,ma2022multimodal,dosovitskiy2021an}, may present a formidable obstacle, producing an overwhelmingly high-dimensional node vector. This will significantly increase computational demands for assessing node similarity.
We argue that this can be addressed by dimension reduction. There are two possible ways: (1) Utilizing model weights of certain layers as node embedding instead of the whole model. (2) Directly learning the low-dimensional node embedding. One simple idea is to leverage Hypernetworks~\citep{shamsian2021personalized} to learn the node embedding with controllable dimensions. 

\section{Future Direction}
\textbf{Handling adversarial clients}. We see the potential of using client topology to tackle adversarial clients. In scenarios where clients, due to system or network failures, disobey the protocols and send arbitrary messages (such as shuffled, sign-flipped, or noised parameters), our client topology learning approach becomes particularly useful. These adversarial behaviors typically result in models that show lower similarity to normal models. By leveraging client topology learning, we can identify these adversarial clients as isolated nodes within the topology.
Once identified, applying centrality measures to these nodes can effectively lower their importance scores. This approach minimizes their impact on the overall model aggregation process. Therefore, we envision that our method holds the potential to tackle adversarial clients. Improving FL’s adversarial robustness is an important and interesting problem, we will leave the exploration of TFL in this direction to future work.